\documentclass[11pt,a4paper]{article}
\usepackage[hyperref]{acl2019}
\usepackage{times}
\usepackage{latexsym}
\usepackage[utf8]{inputenc}
\usepackage[english]{babel}
\usepackage{amsmath, amsthm, amssymb}
\usepackage{graphicx}
\usepackage{xcolor}
\usepackage{todonotes}
\usepackage{soul}
\usepackage{url}
\usepackage{multirow}
\usepackage{booktabs}
\usepackage{flushend}
\usepackage{siunitx}

\aclfinalcopy 

\title{Joint Source-Target Self Attention with Locality Constraints}

\author{José A. R. Fonollosa\hspace{7mm} Noe Casas \hspace{7mm} Marta R. Costa-jussà\\ 
  Universitat Politècnica de Catalunya\\
  \texttt{\{jose.fonollosa,noe.casas,marta.ruiz\}@upc.edu} \\
  \\}

\date{}

\begin{document}
\maketitle
\begin{abstract}

The dominant neural machine translation models are based on
the encoder\textendash decoder structure, and many of them rely on
an unconstrained receptive field over source and target sequences.
In this paper we study a new architecture that breaks with both conventions.
Our simplified architecture consists in the decoder part of
a transformer model, based on self-attention, but with locality
constraints applied on the attention receptive field.

As input for training, both source and target sentences
are fed to the network, which is trained as a language model.
At inference time, the target tokens are predicted autoregressively
starting with the source sequence as previous tokens.

The proposed model achieves a new state of the art of 35.7 BLEU
on IWSLT'14 German-English and matches the best reported results in
the literature on the WMT'14 English-German and WMT'14 
English-French translation benchmarks.
\end{abstract}

\section{Introduction} \label{sec:intro}

In Neural Machine Translation (NMT), the encoder\textendash decoder architectural
pattern has been ubiquitous: all the dominant NMT
models have relied on such an architecture, including
the sequence to sequence model \cite{sutskever2014sequence,cho2014learning},
its variant with attention \cite{bahdanau2014seq2seqattn,luong2015attention},
the Convolutional model \cite{gehring2017conv},
the Transformer model \cite{vaswani2017transformer},
and the Dynamic Convolution model \cite{wu2018dynconv}.

The encoder\textendash decoder architectural pattern consists
of two blocks: the encoder, which receives the source sentence
as input and computes an embedded representation; and
the decoder, which receives the output of the generator and
is trained to generate the target sentence tokens conditioned
also on the target tokens from previous positions.


\newcite{he2018layerwise} proposed an NMT model that does
not have the encoder\textendash decoder separation but learns joint
source-target representations by means of an architecture
that resembles a Transformer Language Model (LM).

In this work we propose to use the idea of joint source-target
representations from \newcite{he2018layerwise} with added
locality constrains \cite{wu2018dynconv} to the receptive field of the self-attention layers \cite{vaswani2017transformer}.

The rest of the article analyzes this proposal following this
structure: section \ref{sec:relatedwork} provides an overview of
related work; section \ref{sec:approach} describes the
proposed approach in detail; section \ref{sec:experiments} describes the
experimental setup, while the obtained results are described in
section \ref{sec:results}; finally, section \ref{sec:conclusion}
draws the final conclusions.

\section{Related Work} \label{sec:relatedwork}

\begin{figure*}[ht!]
\centering
\includegraphics[width=.8\linewidth]{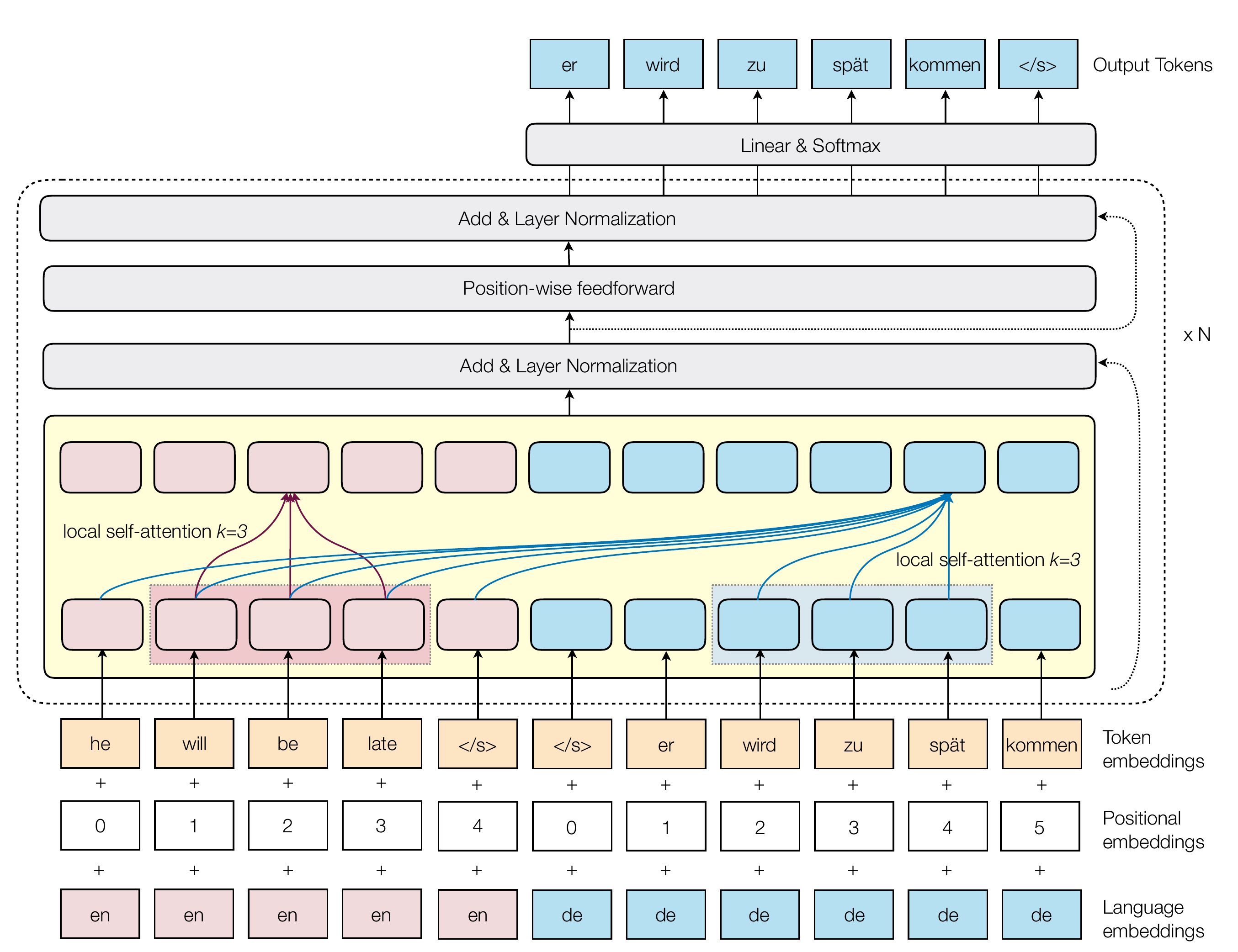}
\caption{Joint source-target architecture with local self-attention.
\label{fig:localselfattn}}
\end{figure*}

The dominant NMT architecture is the Transformer model
proposed by \newcite{vaswani2017transformer}. It consists in an
encoder\textendash decoder architecture, where both encoder and decoder rely
on multi-headed attention blocks. They can be either self-attention
blocks \textemdash if they receive representations of a single
side (source or target) as input\textemdash ~ 
or encoder\textendash decoder attention blocks, where the key and values over
which the attention is computed are the output of the encoder and the
query that drives the weights is an embedded representation of the
target sentence. 

Another recently proposed encoder\textendash decoder architecture that
improves on the results of
the Transformer is the Dynamic Convolution model \cite{wu2018dynconv}.
It makes use of depthwise separable convolutions with kernels with dynamic
weights that are softmax normalized along the temporal dimension.
The width of the kernel (i.e. the receptive field over the temporal
dimension) is progressively increased from the lower
layers to the final ones.
Note that other works also propose to force some notion of
locality into to the attention receptive fields, like
\newcite{yang2018local} who introduce a Gaussian local bias
on the computed self-attention weights.


There have been previous attempts to break from the
encoder\textendash decoder paradigm and combine the purpose of both
elements into a single block:
in the work by \newcite{elbayad2018pervasive}, source and target
sentence embeddings are tiled forming a grid structure over which 2D
masked (causal) convolutions are applied in a stacked manner
to obtain joint source-target representations
from which to derive each next token in the target sequence.

\cite{bapna2018transpattn} also proposed to profit from joint source
and target representations by having the decoder attend to a combination
of the outputs of all encoder layers instead of just the last one.
Without these joint source-target representations the encoder
did not benefit from increasing its number of layers, which suggest
that joint representations can also be key to be able to train
higher capacity models.

Finally, the work by \newcite{he2018layerwise} proposes to use the
decoder part of a Transformer over the concatenation of source and
target sequences, training it as a language model. The attention is
properly masked so that generated target tokens attend to the whole source
sequence and all previous target tokens. Source and target language
embeddings are also added to the source and target token
embedded vectors to help distinguish the two segments.

A similar architecture was also proposed by \cite{lample2019cross} for 
Cross-lingual language model pretraining using pairs of parallel sentences.
However, the proposed Translation Language Model (TLM) is only used 
for cross-lingual classification.

\section{Joint Source-Target Self-Attention with Locality Constraints}
\label{sec:approach}

The studied network architecture is based on the Transformer decoder,
as proposed in \cite{he2018layerwise}. There is
not an independent encoder module and the encoder-decoder attention mechanism is not used.
The architecture consist only of Transformer self-attention blocks.

As shown in figure \ref{fig:localselfattn}, in this architecture we provide
as input to the network the concatenation of source and target sentence
tokens. This design allows the network to learn joint source-target
representations from the early layers.

Given that the processing takes place in batches, we
prepare the source batch and the target batch separately,
applying the appropriate padding in each case,
and then we concatenate the source and target batches in
the sequence dimension.

In order to make the network aware of the two sentences in different languages, 
we follow the approach of \cite{he2018layerwise} and \cite{lample2019cross}.
We add positional embeddings independently to the source and target
(that is, starting the position at $0$ for source and target parts)
as well as language embeddings. Language embeddings are learned 
during training, while for positional embedding we use the
pre-computed sinusoidal variant from \cite{vaswani2017transformer}.

In the normal Transformer self-attention, the receptive field
comprises all tokens for the source sequence and the previous
tokens for the target sequence (i.e. to make it causal).
We propose to apply a reduced receptive field, attending only
to each token's locality.
This is similar to how convolutional kernels are applied
locally in \cite{wu2018dynconv}. The receptive field adopted
grows progressively from the initial layers to the last ones.

In order to implement such locally constrained attention, a
masking approach is followed, similar to the causal masking
normally used in the decoder of the Transformer model. In our
case, the mask forms a \textit{band}
with the specified receptive field size as width. At the source
side, the mask is centered at each token while, at the target
side, the mask includes only the previous tokens in order
to keep causality.

The loss is defined as a standard categorical cross-entropy
applied only to the target output tokens, with the usual label 
smoothing with 0.1 weight for the uniform prior distribution over the vocabulary
of Transformer-based architectures.

At inference time, the model works like any standard autoregressive LM,
but in this case we use the source sentence tokens as starting
point for the next token generation.

\section{Experimental Setup} \label{sec:experiments}

\begin{table*}[t]
\begin{center}
\caption{Translation quality evaluation (BLEU scores). \label{tab:results}}
\begin{tabular}{lccc}
\toprule
\multirow{2}{*}{\vspace{-2mm}Model} & \multicolumn{2}{c}{WMT'14} & IWSLT'14 \\
\cmidrule(l{2pt}r{2pt}){2-3} \cmidrule(l{2pt}r{2pt}){4-4}
 & EN-DE & EN-FR & DE-EN  \\
\hline
\rule{0pt}{2.0ex}\newcite{vaswani2017transformer} & 28.4 & 41.0 & 34.4 \\
\newcite{ahmed2018weighted} & 28.9 & 41.4 & - \\
\newcite{chen2018combining} & 28.5 & 41.0 & - \\
\newcite{shaw2018relative} & 29.2 & 41.5 & - \\
\newcite{ott2018scaling} & 29.3 & 43.2 & - \\
\newcite{wu2018dynconv} & \textbf{29.7} & 43.2 & 35.2 \\
\newcite{he2018layerwise} & 29.0 & - & 35.1 \\
\specialrule{1pt}{-1pt}{0pt}
Joint Self-attention& \textbf{29.7} & 43.2 & 35.3 \\
Local Joint Self-attention & \textbf{29.7} & \textbf{43.3} & \textbf{35.7} \\
\bottomrule
\end{tabular}
\end{center}
\end{table*}

In order to assess the translation quality of the proposed
architectures experimentally and make the evaluation comparable to previous work, we ran
experiments on standard benchmark datasets with the usual setups and evaluation 
protocols, including IWSLT'14 German-English, WMT'14 English-German and
WMT'14 English-French.

For WMT'14 en-de, we use for training the preprocessed data released
by \newcite{vaswani2017transformer} as part of \texttt{tensor2tensor},
which actually contains 4.5M sentence pairs from the WMT'16 training data, 
tokenized and byte-pair encoded (BPE) \citep{sennrich2016bpe}
with a joint source and target vocabulary of 32K tokens, while for evaluation
we use newstest2014.

For WMT'14 en-fr, we use the preprocessing script \texttt{prepare-wmt14en2fr.sh}
included with the \texttt{fairseq} library%
\footnote{\url{https://github.com/pytorch/fairseq}}%
, which tokenizes, cleans
and normalizes punctuation
with the utilities from Moses \cite{moses}, and computes a joint source target BPE
vocabulary of 40K tokens.

For IWSLT'14 de-en, we use the analogous \texttt{fairseq} script
\texttt{prepare-iwslt14.sh}, but we extended the vocabulary to 31K joint
source target BPE  tokens as \newcite{he2018layerwise}. The script
provides the usual lowercased data with 160K training sentence pairs.

In order to evaluate the translation quality, we use case-sensitive tokenized
BLEU scores \cite{papineni2002bleu}, for WMT'14 en-de and WMT'14 en-fr.
For WMT'14 en-de, we also performed compound splitting,
like the works we are comparing to. For the computation of BLEU itself, we used
\texttt{sacrebleu} \cite{post2018sacrebleu}. For the IWSLT'14 de-en we also
present comparable tokenized BLEU results.

The hyperparameter configuration used for our experiments with the joint source-target
self-attention for the IWSLT'14 de-en benchmark consists on 14 layers, with an embedding 
size of 256, feedforward expansion size of 1024 and 4 attention heads. For the version 
with locality constraints, the attention window sizes from input layers to output layers are
3, 5, 7, 9, 11, 13, 15, 17, 21, 25, 29, 33, 37, 41.

The configuration for WMT'14 en-de and en-fr also has 14 layers and
the same hyperparameter values as the \texttt{transformer-big}
setup from \cite{vaswani2017transformer}: an embedding 
size of 1024, feedforward expansion size of 4096 and 16 attention heads.
The attention window sizes, from input layers to output layers are
7, 15, 31, 63, 63, 63, 63, 63, 63, 63, 63, 63, 63, 63.

In all configurations, the number of layers were chosen to
have approximately the same number of total trainable parameters
as the big Transformer model configuration used for each dataset.

The proposed NMT architecture was implemented on top of the \texttt{fairseq}
library. The training parameters are based on \cite{wu2018dynconv}: Adam optimizer, 
batches of 500K source tokens for WMT benchmarks and 4K for IWSLT de-en, 
30K training steps for WMT en-de, 80K steps for WMT en-fr and 85K for IWSLT de-en.
The learning rate is linearly warmed for the first 10K steps up to a maximum of 
\num{e-3} for IWSLT and \num{0.5e-3} for WMT benchmarks,
followed by an inverse square root scheduler on IWSLT and a cosine rate with a single cycle on WMT.
The source code to reproduce our results and pretrained models are available
at \url{https://github.com/jarfo/joint}

\section{Results} \label{sec:results}

Table \ref{tab:results} presents a comparison of
the translation quality measured via BLEU score between
the currently dominant Transformer \cite{vaswani2017transformer}
and Dynamic Convolutions \cite{wu2018dynconv} models,
as well as the work by \newcite{he2018layerwise}, which
also proposes a joint encoder-decoder structure, and
also other refinements over the transformer architecture
like
\cite{ahmed2018weighted},
\cite{chen2018combining},
\cite{shaw2018relative}
and \cite{ott2018scaling} .

The entry \textit{Joint Self-attention} corresponds to the results 
of our implementation of \cite{he2018layerwise}, that significantly improves
the original results by 0.7 BLEU point on the WMT14 de-en benchmark, and 0.2 on IWSLT.
The same architecture with the proposed locality constraints (\textit{Local Joint Self-attention}) 
establishes a new state of the art in IWSLT'14 de-en with 35.7 BLEU,
surpassing all previous published results by at least in 0.5 BLEU, and our results with 
the unconstrained version by 0.4. 

The \textit{Joint Self-attention} model obtains the same SoTA BLEU score of \cite{wu2018dynconv} on WMT'14 en-de, and the same SoTA score of \cite{ott2018scaling} and \cite{wu2018dynconv} on WMT'14 en-fr. The local attention constraints do not provide a significant gain on these bigger models, but it improves the BLEU score on WMT'14 en-fr to a new SoTA of $43.3$.

\section{Conclusion} \label{sec:conclusion}

In this work we studied an NMT architecture that merges the
classical encoder-decoder components into a single block
that learns joint source-target representations starting from its
initial layers and makes use of locality constraints over
the attention receptive field.

Our experiments show that the joint source-target model
with local attention achieve state of the art results on
standard WMT benchmarks, and significantly improves the 
best published result on the IWSLT'14 de-en benchmark.

\ifaclfinal
\section*{Acknowledgments}

This work is partially supported by Lucy Software / United Language Group (ULG)
and the Catalan Agency for Management of University and Research Grants (AGAUR)
through an Industrial PhD Grant.
This work is also supported in part by the
Spanish Ministerio de Economía y Competitividad,
the European Regional Development Fund
and the Agencia Estatal de Investigación,
through the postdoctoral senior grant Ramón y Cajal, contract TEC2015-69266-P
(MINECO/FEDER,EU) and contract PCIN-2017-079 (AEI/MINECO).

\fi


\bibliography{biblio}

\begin{thebibliography}{20}
\expandafter\ifx\csname natexlab\endcsname\relax\def\natexlab#1{#1}\fi

\bibitem[{Ahmed et~al.(2017)Ahmed, Keskar, and Socher}]{ahmed2018weighted}
Karim Ahmed, Nitish~Shirish Keskar, and Richard Socher. 2017.
\newblock \href {https://arxiv.org/abs/1711.02132} {Weighted transformer
  network for machine translation}.
\newblock \emph{arXiv preprint arXiv:1711.02132}.

\bibitem[{Bahdanau et~al.(2014)Bahdanau, Cho, and
  Bengio}]{bahdanau2014seq2seqattn}
Dzmitry Bahdanau, Kyunghyun Cho, and Yoshua Bengio. 2014.
\newblock \href {https://arxiv.org/abs/1409.0473} {Neural machine translation
  by jointly learning to align and translate}.
\newblock \emph{arXiv preprint arXiv:1409.0473}.

\bibitem[{Bapna et~al.(2018)Bapna, Chen, Firat, Cao, and
  Wu}]{bapna2018transpattn}
Ankur Bapna, Mia Chen, Orhan Firat, Yuan Cao, and Yonghui Wu. 2018.
\newblock \href {http://aclweb.org/anthology/D18-1338} {Training deeper neural
  machine translation models with transparent attention}.
\newblock In \emph{Proceedings of the 2018 Conference on Empirical Methods in
  Natural Language Processing}, pages 3028--3033. Association for Computational
  Linguistics.

\bibitem[{Chen et~al.(2018)Chen, Firat, Bapna, Johnson, Macherey, Foster,
  Jones, Schuster, Shazeer, Parmar, Vaswani, Uszkoreit, Kaiser, Chen, Wu, and
  Hughes}]{chen2018combining}
Mia~Xu Chen, Orhan Firat, Ankur Bapna, Melvin Johnson, Wolfgang Macherey,
  George Foster, Llion Jones, Mike Schuster, Noam Shazeer, Niki Parmar, Ashish
  Vaswani, Jakob Uszkoreit, Lukasz Kaiser, Zhifeng Chen, Yonghui Wu, and
  Macduff Hughes. 2018.
\newblock \href {http://aclweb.org/anthology/P18-1008} {The best of both
  worlds: Combining recent advances in neural machine translation}.
\newblock In \emph{Proceedings of the 56th Annual Meeting of the Association
  for Computational Linguistics (Volume 1: Long Papers)}, pages 76--86.
  Association for Computational Linguistics.

\bibitem[{Cho et~al.(2014)Cho, van Merrienboer, Gulcehre, Bahdanau, Bougares,
  Schwenk, and Bengio}]{cho2014learning}
Kyunghyun Cho, Bart van Merrienboer, Caglar Gulcehre, Dzmitry Bahdanau, Fethi
  Bougares, Holger Schwenk, and Yoshua Bengio. 2014.
\newblock \href {https://doi.org/10.3115/v1/D14-1179} {Learning phrase
  representations using rnn encoder--decoder for statistical machine
  translation}.
\newblock In \emph{Proceedings of the 2014 Conference on Empirical Methods in
  Natural Language Processing (EMNLP)}, pages 1724--1734. Association for
  Computational Linguistics.

\bibitem[{Elbayad et~al.(2018)Elbayad, Besacier, and
  Verbeek}]{elbayad2018pervasive}
Maha Elbayad, Laurent Besacier, and Jakob Verbeek. 2018.
\newblock \href {http://aclweb.org/anthology/K18-1010} {Pervasive attention:
  {2D} convolutional neural networks for sequence-to-sequence prediction}.
\newblock In \emph{Proceedings of the 22nd Conference on Computational Natural
  Language Learning}, pages 97--107. Association for Computational Linguistics.

\bibitem[{Gehring et~al.(2017)Gehring, Auli, Grangier, Yarats, and
  Dauphin}]{gehring2017conv}
Jonas Gehring, Michael Auli, David Grangier, Denis Yarats, and Yann~N. Dauphin.
  2017.
\newblock \href {http://proceedings.mlr.press/v70/gehring17a.html}
  {Convolutional sequence to sequence learning}.
\newblock In \emph{Proceedings of the 34th International Conference on Machine
  Learning, {ICML} 2017, Sydney, NSW, Australia, 6-11 August 2017}, pages
  1243--1252.

\bibitem[{He et~al.(2018)He, Tan, Xia, He, Qin, Chen, and
  Liu}]{he2018layerwise}
Tianyu He, Xu~Tan, Yingce Xia, Di~He, Tao Qin, Zhibo Chen, and Tie-Yan Liu.
  2018.
\newblock \href
  {http://papers.nips.cc/paper/8019-layer-wise-coordination-between-encoder-and-decoder-for-neural-machine-translation.pdf}
  {Layer-wise coordination between encoder and decoder for neural machine
  translation}.
\newblock In S.~Bengio, H.~Wallach, H.~Larochelle, K.~Grauman, N.~Cesa-Bianchi,
  and R.~Garnett, editors, \emph{Advances in Neural Information Processing
  Systems 31}, pages 7955--7965. Curran Associates, Inc.

\bibitem[{Koehn et~al.(2007)Koehn, Hoang, Birch, Callison-Burch, Federico,
  Bertoldi, Cowan, Shen, Moran, Zens, Dyer, Bojar, Constantin, and
  Herbst}]{moses}
Philipp Koehn, Hieu Hoang, Alexandra Birch, Chris Callison-Burch, Marcello
  Federico, Nicola Bertoldi, Brooke Cowan, Wade Shen, Christine Moran, Richard
  Zens, Chris Dyer, Ond\v{r}ej Bojar, Alexandra Constantin, and Evan Herbst.
  2007.
\newblock \href {https://aclanthology.info/papers/P07-2045/p07-2045} {Moses:
  Open source toolkit for statistical machine translation}.
\newblock In \emph{Proceedings of the 45th Annual Meeting of the ACL on
  Interactive Poster and Demonstration Sessions}, ACL '07, pages 177--180.
  Association for Computational Linguistics.

\bibitem[{Lample and Conneau(2019)}]{lample2019cross}
Guillaume Lample and Alexis Conneau. 2019.
\newblock \href {https://arxiv.org/abs/1901.07291} {Cross-lingual language
  model pretraining}.
\newblock \emph{arXiv preprint arXiv:1901.07291}.

\bibitem[{Luong et~al.(2015)Luong, Pham, and Manning}]{luong2015attention}
Thang Luong, Hieu Pham, and Christopher~D. Manning. 2015.
\newblock \href {https://doi.org/10.18653/v1/D15-1166} {Effective approaches to
  attention-based neural machine translation}.
\newblock In \emph{Proceedings of the 2015 Conference on Empirical Methods in
  Natural Language Processing}, pages 1412--1421. Association for Computational
  Linguistics.

\bibitem[{Ott et~al.(2018)Ott, Edunov, Grangier, and Auli}]{ott2018scaling}
Myle Ott, Sergey Edunov, David Grangier, and Michael Auli. 2018.
\newblock \href {http://www.aclweb.org/anthology/W18-64001} {Scaling neural
  machine translation}.
\newblock In \emph{Proceedings of the Third Conference on Machine Translation},
  pages 1--9, Belgium, Brussels. Association for Computational Linguistics.

\bibitem[{Papineni et~al.(2002)Papineni, Roukos, Ward, and
  Zhu}]{papineni2002bleu}
Kishore Papineni, Salim Roukos, Todd Ward, and Wei-Jing Zhu. 2002.
\newblock \href {https://aclanthology.info/papers/P02-1040/p02-1040} {{BLEU}: a
  method for automatic evaluation of machine translation}.
\newblock In \emph{Proceedings of the 40th annual meeting on association for
  computational linguistics}, pages 311--318. Association for Computational
  Linguistics.

\bibitem[{Post(2018)}]{post2018sacrebleu}
Matt Post. 2018.
\newblock \href {http://aclweb.org/anthology/W18-6319} {A call for clarity in
  reporting {BLEU} scores}.
\newblock In \emph{Proceedings of the Third Conference on Machine Translation:
  Research Papers}, pages 186--191. Association for Computational Linguistics.

\bibitem[{Sennrich et~al.(2016)Sennrich, Haddow, and Birch}]{sennrich2016bpe}
Rico Sennrich, Barry Haddow, and Alexandra Birch. 2016.
\newblock \href {https://aclanthology.info/papers/P16-1162/p16-1162} {Neural
  machine translation of rare words with subword units}.
\newblock In \emph{Proceedings of the 54th Annual Meeting of the Association
  for Computational Linguistics, {ACL} 2016, August 7-12, 2016, Berlin,
  Germany, Volume 1: Long Papers}.

\bibitem[{Shaw et~al.(2018)Shaw, Uszkoreit, and Vaswani}]{shaw2018relative}
Peter Shaw, Jakob Uszkoreit, and Ashish Vaswani. 2018.
\newblock \href {https://doi.org/10.18653/v1/N18-2074} {Self-attention with
  relative position representations}.
\newblock In \emph{Proceedings of the 2018 Conference of the North American
  Chapter of the Association for Computational Linguistics: Human Language
  Technologies, Volume 2 (Short Papers)}, pages 464--468. Association for
  Computational Linguistics.

\bibitem[{Sutskever et~al.(2014)Sutskever, Vinyals, and
  Le}]{sutskever2014sequence}
Ilya Sutskever, Oriol Vinyals, and Quoc~V Le. 2014.
\newblock \href
  {https://papers.nips.cc/paper/5346-sequence-to-sequence-learning-with-neural-networks.pdf}
  {Sequence to sequence learning with neural networks}.
\newblock In \emph{Advances in neural information processing systems}, pages
  3104--3112.

\bibitem[{Vaswani et~al.(2017)Vaswani, Shazeer, Parmar, Uszkoreit, Jones,
  Gomez, Kaiser, and Polosukhin}]{vaswani2017transformer}
Ashish Vaswani, Noam Shazeer, Niki Parmar, Jakob Uszkoreit, Llion Jones,
  Aidan~N Gomez, {\L}ukasz Kaiser, and Illia Polosukhin. 2017.
\newblock Attention is all you need.
\newblock In I.~Guyon, U.~V. Luxburg, S.~Bengio, H.~Wallach, R.~Fergus,
  S.~Vishwanathan, and R.~Garnett, editors, \emph{Advances in Neural
  Information Processing Systems 30}, pages 6000--6010. Curran Associates, Inc.

\bibitem[{Wu et~al.(2019)Wu, Fan, Baevski, Dauphin, and Auli}]{wu2018dynconv}
Felix Wu, Angela Fan, Alexei Baevski, Yann Dauphin, and Michael Auli. 2019.
\newblock \href {https://openreview.net/forum?id=SkVhlh09tX} {Pay less
  attention with lightweight and dynamic convolutions}.
\newblock In \emph{International Conference on Learning Representations}.

\bibitem[{Yang et~al.(2018)Yang, Tu, Wong, Meng, Chao, and
  Zhang}]{yang2018local}
Baosong Yang, Zhaopeng Tu, Derek~F. Wong, Fandong Meng, Lidia~S. Chao, and Tong
  Zhang. 2018.
\newblock \href {http://aclweb.org/anthology/D18-1475} {Modeling localness for
  self-attention networks}.
\newblock In \emph{Proceedings of the 2018 Conference on Empirical Methods in
  Natural Language Processing}, pages 4449--4458. Association for Computational
  Linguistics.

\end{thebibliography}
\bibliographystyle{acl_natbib}

\end{document}